# Multilingual Artificial Datasets for AI-Driven Cardiology: the CliTe and ParaCliTe corpora

Lorenzo Pratesi (1)*, Eulalia Farre-Maduell (2)*, Machteld Boonstra (3)*, Michele Miranda (1, 4), Jonas Mack (1), Eduard Rodenas-Alesina (6, 7, 8), Jonathan Kossoff (9), Corinne Allaart (10), Sébastien Bratières (1), Manuela-Daniela Danu (11,17), Isabella Fumarulo (5), Milou Gamage (9), Antonio Iaconelli (5, 12), Pavel Leinveber (13), Felix Lindberg (14), Pavel Macháč (13), Vladimíra Macíková (13), Gianluigi Savarese (14), Raffaele Scorza (5), Bram van Es (10), Emma Weiss (15, 16), Andrea Guala (6, 8), Martin Krallinger (2), and the DataTools4Heart NLP group.

**Affiliations**

1 Translated SRL, Rome, Italy
2 Barcelona Supercomputing Center, Barcelona, Spain
3 Amsterdam University Medical Center, University of Amsterdam, Amsterdam, The Netherlands
4 Sapienza University of Rome, Rome, Italy
5 Department of Cardiovascular Medicine, Fondazione Policlinico Universitario A. Gemelli IRCCS, Rome, Italy.
6 Vall d'Hebron Research Institute, Barcelona, Spain
7 Vall d'Hebron University Hospital, Barcelona, Spain
8 Biomedical Research Networking Center on Cardiovascular Diseases, Instituto de Salud Carlos III, Madrid, Spain
9 University College London Hospital, London, UK
10 University Medical Center Utrecht, Utrecht University, Utrecht, the Netherlands
11 Siemens SRL, Brasov, Romania
12 School of Cardiovascular and Metabolic Health, College of Medical, Veterinary and Life Sciences, University of Glasgow, Glasgow, UK.
13 International Clinical Research Center, St. Anne's University Hospital Brno, Brno, Czech Republic
14 Department of Clinical Science and Education, Södersjukhuset; Karolinska Institutet, Stockholm, Sweden
15 Colentina Hospital, Bucharest, Romania
16 Carol Davila University of Medicine and Pharmacy, Bucharest, Romania
17 Automation and Information Technology, Transilvania University of Brasov, Brasov, Romania.

* These Authors contributed equally and are thus co-first Authors

**Word count**:4189

**Corresponding Authors:** Andrea Guala (andrea.guala@vhir.org, Vall d'Hebron Research Institute, Barcelona, Spain) and Martin Krallinger (krallinger.martin@gmail.com, Barcelona Supercomputing Center, Barcelona, Spain)

**Abbreviations**

AI = Artificial Intelligence; MT = machine translation; NLP = Natural Language Processing

PLP = perplexity

# Abstract

**Background and Aims**

Comparable, multilingual, validated cardiology corpora are critical to advance trustworthy natural language processing (NLP) models for cardiology but such resources are scarce due to privacy concerns. This study aimed to cover this gap by presenting suitable artificial corpus creation guidelines and by releasing 3 multilanguage corpora.

**Methods**

Following purpose-specific publicly-available corpus construction guidelines, cardiologists from 7 European hospitals generated artificial clinical records in Czech, Dutch, English, Italian, Romanian, Spanish and Swedish, ensuring clinical diversity and sex balance, resulting in the CliTe corpus. A CliTe subset underwent a multi-stage manual translation into the other 6 languages, resulting in the parallel ParaCliTe corpus, while the remaining reports were automatically translated, constituting ParaCliTe Silver.

**Results**

CliTe is a collection of 699 artificial cardiology reports, mostly outpatient notes (340 documents), discharge reports (263), and inpatient notes (89), describing fictional patients with ischaemic heart disease (199 patients), heart failure (188), and valvular disease (79), among others. ParaCliTe contains reports from 94 fictitious patients, each available in seven languages with sentence-level alignment, resulting in 658 documents. The corpus mainly comprises discharge reports (322 documents), and outpatient (196) and inpatient (140) notes, describing patients with heart failure (399 documents), ischaemic heart disease (287), and valvular disease (238). Empirical results illustrate the utility of ParaCliTe and highlight the limitations of state-of-the-art NLP models adapted to the cardiology domain.

**Conclusions**

The creation of multilingual, equitable and realistic corpora of artificial patients capable of strengthening NLP performance for cardiology is feasible when supported by detailed guidelines.

# Introduction

Cardiovascular diseases are the leading cause of death globally [1], and socioeconomically disadvantaged people are at higher risk [2]. The rapid advancement of artificial intelligence (AI) solutions applied to cardiology, particularly deep learning, have demonstrated the ability to identify information that is imperceptible to humans [3]–[5], uncover hidden long-term risks in low-risk individuals [6], [7], assist clinical documentation and coding [8], and support clinical decision-making [9], [10]. Among the various types of data used in cardiology, unstructured medical text remains underexploited, despite being the data type where detailed patient information such as symptoms, lifestyle and social determinants of health as well as clinical considerations of physicians can mostly be found [8]. As a considerable fraction of all clinical information is stored as unstructured free text, exploiting free texts it is therefore important for a holistic view of the patient status [11].

To comply with privacy legislation while permitting the reuse of reports for research purposes, exhaustive automatic removal of information that might reidentify patients and healthcare professionals is required. Despite advanced technical solutions for medical text anonymization and de-identification, this process remains particularly challenging [12]. Consequently, only a limited number of de-identified clinical record datasets have been released to the public. These datasets are limited in size and extensibility potential [13], and generally restricted to a single language, mainly English, or only partially bilingual [14]–[16]. However, clinical cardiology is practiced and recorded across Europe in local languages, which further calls for the creation of interoperable, international datasets, a critical resource for the development and benchmarking of Natural Language Processing (NLP) systems. In this context, the availability of comparable multilingual cardiology datasets spanning multiple healthcare institutions across different countries is critical to implement transferable AI research innovations into real-world cardiology practice.

Beyond linguistic differences, implicit variations in patient populations and data collection practices can embed population and health system characteristics into the data, which could limit generalizability and introduce bias into trained NLP models. Although advanced machine translation (MT) methods could help reduce this gap [17], [18], their accuracy for low-resource languages remains largely unvalidated, with few translation tools meeting stringent standards, thus requiring human editing to ensure reliability [19], while local patient variations are not captured. Currently, most Large Language Models (LLMs) exhibit a strong bias toward English [20], suggesting the risk of substantial inequities that could impact LLM-enhanced care. Thus, there is a pressing need for multilingual resources to facilitate the development of effective comparable pan-European clinical language models [21], which is specifically recognized as a requirement to realize the European Health Data Space. Novel strategies for synthetic data generation based on local clinical practice may offer a pragmatic way to address both privacy constraints and multilingual scarcity [22], [23], but concerns related to how these texts closely mirror real-world clinical practices and the possibility that they reinforce LLMs biases still remain.

In response to the aforementioned limitations, and in the context of the Horizon Europe-funded projects DataTools4Heart and AI4HF, we present and openly release a set of guidelines to foster the unbiased, scalable and accurate generation of artificial texts suitable for the training, fine tuning and validation of multilingual NLP systems. We further openly release the first three multilingual cardiology corpora: CliTe (Clinical Text Corpus), which contains 699 case reports in 7 European languages, ParaCliTe, a subset of CliTe that underwent a rigorous multi-stage professional translation and validation, and ParaCliTe Silver, an extensive collection of multilingual translation of CliTe texts obtained using Machine Translation (MT).

## Methods

This section includes the methodology followed for the development of guidelines for the unbiased generation of artificial case reports in multiple languages and for the creation of the multilingual cardiology corpora CliTe, ParaCliTe and ParaCliTe Silver. A visual overview of the key steps is reported in Figure 1, while CliTe, ParaCliTe and ParaCliTe Silver can be accessed at the Zenodo repository (https://zenodo.org/records/22641301).

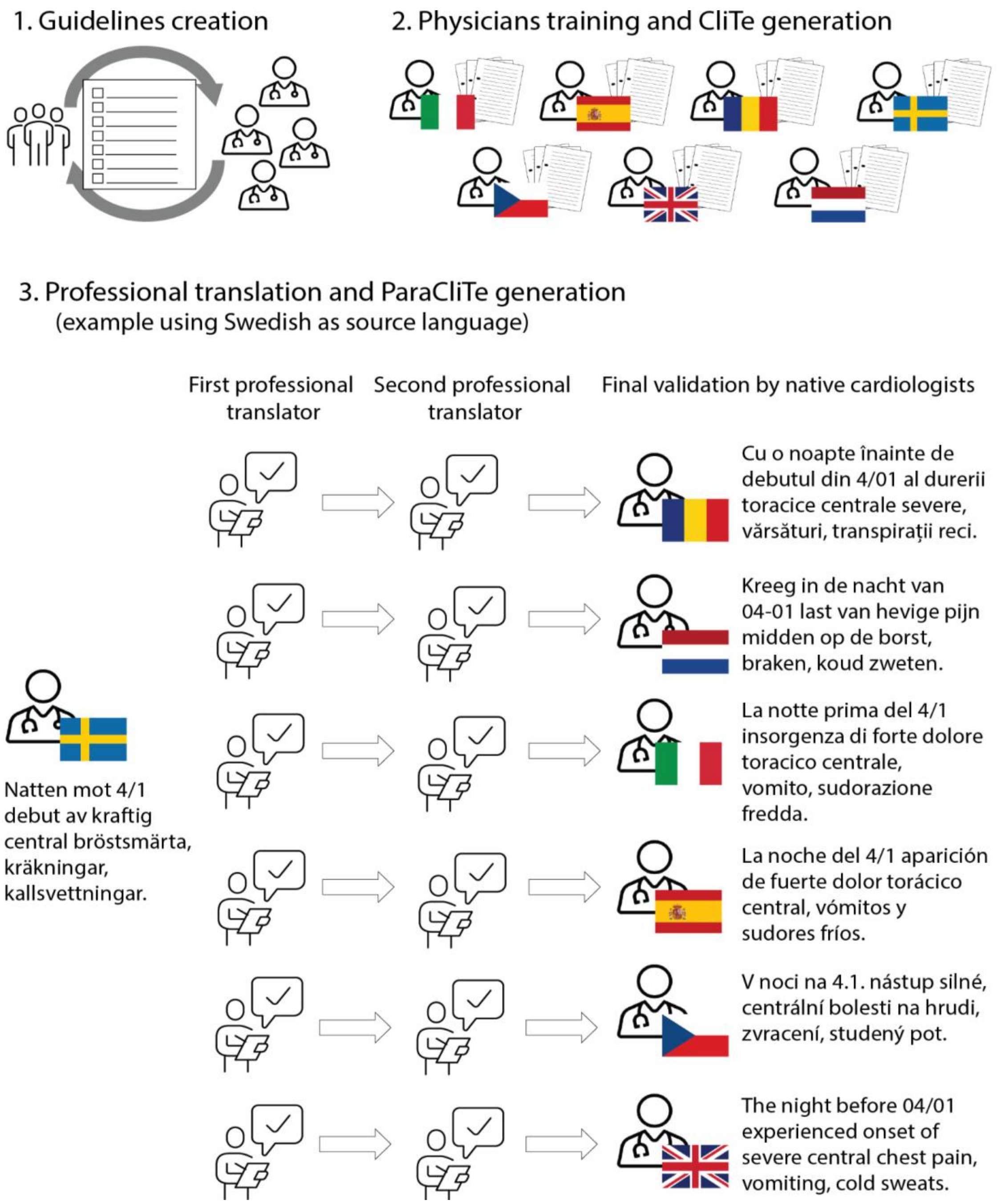

**Figure 1. Representation of the multilingual, multi-step translation process.** The top row represents the process of development of the Guidelines and the subsequent generation of CliTe corpus. The bottom row showcases an example of the translation process, where a fictional medical case is written in Swedish by a participating clinician. The text is then translated in each of the other six languages by a first translator per language, followed by a first revision by another professional translator, and a final review by a native clinician.

## Guidelines for the generation of artificial clinical texts

Crafting credible and consistent artificial reports in medicine demands tailored guidelines to ensure the data align with the structure of the target clinical documents while being clinically realistic, ethically robust, and properly documented. Clear data construction principles also ensure reproducibility and model generalizability in real-world healthcare settings. By standardizing the whole data creation process, these guidelines enhance transparency and mitigate risks when AI models trained on artificial data are deployed in clinical care.

The initial draft of the guidelines defined the overall framework and steps for the creation of the corpus. This included specifications about the document format, file-naming conventions, and training procedures for contributors. Next, two seed reports drafted by a cardiologist in Spanish and translated into each of the six remaining target languages were provided as initial examples. Subsequently, one or more cardiologists from each of the seven participating hospitals created an initial sample in local language and provided feedback on the guidelines to improve them.

To build a realistic corpus replicating local hospital documents as accurately as possible, it is essential that these artificial records avoid harmful biases and reflect the true diversity of patients, encompassing sex, ethnicity, age, and a wide range of cardiovascular conditions. It is also crucial to include some of the real-world habits and imperfections present in clinical records, often written under severe time constraints, such as ad hoc acronyms and abbreviations, and typographical errors. Therefore, these biases were explicitly addressed in the guidelines to ensure balanced representation of population characteristics.

The *Guiding Principles for Multilingual Generation of Artificial Clinical Reports* are published in the Zenodo repository (https://zenodo.org/records/21220631).

## Creation of the CliTe corpus

The guidelines were followed to build the CliTe dataset, a textual corpus consisting of a collection of approximately 100 reports narrating fictionalized medical cases per each of the 7 languages, totalling 699 original artificial clinical documents. These reports were produced by clinical professionals representing the expertise from the following European hospitals and languages of the DataTools4Heart project: Saint Anne's University Hospital in Brno for Czech, Utrecht University Medical Centre and Amsterdam University Medical Centre for Dutch, University College London Hospital for English, Fondazione Policlinico A. Gemelli in Rome for Italian, Bucharest Emergency Clinical Hospital for Romanian, Vall d'Hebron University Hospital in Barcelona for Spanish, and Karolinska University Hospital in Stockholm for Swedish. Note that no ethical approval was sought regarding the creation of these cases, since no personal data were involved.

After an initial sample of artificial cardiology cases was created by one or more clinical experts from each site and later verified by the NLP experts, clinicians were encouraged to generate a large and diverse sample of clinical reports. Length of documents varies widely, reflecting local practices and considering that some clinicians felt longer reports (such as discharge letters) could be more relevant than shorter reports (as outpatient visit notes), and vice-versa. To enhance post-production uses of CliTe, the generated text had to be accompanied by relevant metadata. Thus, authors were instructed to label each report with the original language, patient sex, age, and primary cardiac diagnosis and type of document (admission notes, outpatient reports, discharge summaries and referral letters).

## Creation of the ParaCliTe corpus

A total of 94 CliTe reports were subsequently selected to undergo rigorous multi-stage professional translations into the remaining six languages, as visually summarised in Figure 1, resulting in a parallel corpus called ParaCliTe. The selection of documents aimed at balancing sex in both the number of reports and their total word count per language, while reflecting the type of document that the local clinicians considered more relevant. In this respect, some languages included larger hospital discharge reports (Czech, Romanian, Spanish), while others contained shorter in- and out-patient notes (Dutch, Swedish, English). The selection resulted in a corpus that contains a minimum of 7,000 words of original text per language and sex parity, with 52% of documents (24,717 words) referring to female patients, and 48% of documents (26,670 words) to male patients.

After this selection process, professional medical translators were recruited for each of the 42 language pairs, with at least two translators assigned per direction to ensure quality control. The translators were selected based on previous experience in medical translation and were informed of the project's aim. They were instructed to treat the simulated reports as actual clinical texts and to preserve the original style, abbreviations and structure. Instructions were given to log ambiguous acronyms, domain-specific terms and abbreviations as well as any unclear part of the text, which were later clarified by clinicians in a final review round.

The translation process involved multiple stages and was supported by MateCat [24], an open-source computer-assisted translation platform that enables real-time editing, alignment, and terminology consistency checks. The process started with a first-pass translation by a professional translator, which was followed by a quality review by a second independent professional translator. A final feedback loop between translators and clinicians was performed to clarify domain-specific terms and abbreviations. For some language directions, such as Romanian to Czech, qualified direct translators were not available, requiring translations to be completed via English as the pivot language. Specifically, this process involved the use of a previously translated English version of a given non-English text, which was used as a starting point for the following translation to the target language. Quality assurance combined technical and expert review components, while MateCat's intrinsic functions provided validation of segmentation, tag consistency, and completeness. Human revision addressed linguistic adequacy, terminological precision, and stylistic coherence.

After the generation of all translations, each text was further categorized at complete report level by the members of the coordinating team to ensure consistency. During this annotation process, the following metadata were added: presence of relevant comorbidities and risk factors (smoking status, reduced mobility, obesity or overweight, diabetes, hypertension, kidney disease and chronic obstructive pulmonary disease), presence of symptoms at presentation (shortness of breath, oedema, chest pain, palpitation and New York Heart Association -NYHA- class), presence of arrhythmias (including its categorization into atrial fibrillation or others), and presence of implanted devices (pacemakers, defibrillators, cardiac resynchronization therapies, artificial heart valves, coronary stents, and left ventricular assist devices) (see Supplementary Material).

The final corpus consists of artificial reports featuring fictional patients aligned with their six corresponding translations and linked metadata to allow for seamless linkage to each corresponding source document at document level. A finer segmentation of the text was provided with the MateCat segmentation tool, resulting in the segmentation of each report into sentences, which were matched to its six translations. All the documents were normalized into a plain text (.txt) format. ParaCliTe also includes a clarifications file that records ambiguous or underspecified expressions for which translators requested clarification from the clinicians during the translation workflow, together with the resolved interpretation, key information to build a cross-lingual cardiology glossary grounded in real translation-time uncertainties. Finally, following standard NLP training protocols and to foster transparent downstream NLP applications, the dataset was partitioned into training, validation, and test sets using a 64%, 16%, and 20% split, respectively.

## Demonstration of ParaCliTe and ParaCliTe Silver utility

We hypothesized that current off-the-shelf MT and more generally language models would struggle to capture the linguistic aspects and domain-specific nuances present in ParaCliTe, and that fine-tuning these models on ParaCliTe could improve the quality of automatic translations. To substantiate these claims, an experiment was designed to measure performance changes before and after fine-tuning on the ParaCliTe training splits of two distinct paradigms of neural architectures, namely MT models and LLMs.

We evaluated two state-of-the-art MT models (NLLB-600M-distilled and NLLB-3B [25]). Zero-shot and fine-tuned performance were directly compared on the ParaCliTe test set using standard generative metrics (COMET [26], COMET-Kiwi [27], BLEU [28], chrF++ [29]). We evaluated six state-of-the-art models: Llama 3.2 1B, Llama 3.1 8B [30], Mistral v0.3 7B [31], Qwen3 8B [32], Gemma 2 9B [33], and Mistral-Nemo-12B [34]. Due to formatting inconsistencies in raw zero-shot LLM outputs, we bypassed generative decoding and intrinsically evaluated the baseline models by measuring the perplexity (PPL) of ground-truth target translations given the source text. Conversely, as the fine-tuning resolved output formatting constraints, we were then able to evaluate the fine-tuned LLMs using the exact same generative metrics applied to the MT models, effectively establishing a unified benchmark. Comprehensive details regarding the experimental setup, fine-tuning procedures, hyperparameters, and computational resources are provided in the Supplementary Materials.

Finally, to expand the dataset's utility, the 605 remaining untranslated CliTe documents were automatically translated to create the ParaCliTe Silver corpus. This pipeline utilized the MT model which demonstrated the highest domain adaptation and translation quality. The post-processing pipeline involved segmenting the documents at the sentence level using MateCat, translating each sentence into the six target languages, reconstructing the complete documents by concatenating the sentences, and finally normalizing them into plain text to ensure similar formatting as ParaCliTe. To maximize downstream utility, the corpus is released in two complementary formats: normalized, document-level files and parallel, segment-level files.

# Results

## CliTe

Based on the instructions and criteria defined in the clinical corpus construction guidelines, a total of 699 artificial case reports of fictional adult cardiology patients were created (Table 1). The reports were evenly distributed across languages, with 100 reports in Czech, Romanian, English, Dutch, Spanish, and Swedish, and 99 in Italian.

**Table 1**: Overview of the characteristics of CliTe, ParaCliTe and ParaCliTe silver corpora.

| | **CliTe** | **ParaCliTe** | **ParaCliTe Silver** |
|---|---|---|---|
| **Dataset characteristics** | | | |
| Total word count (source only; including translations) | 430,455; - | 51,079; 394,447 | 379,376; 3,141,659 |
| Total number of documents | 699 | 658 | 4,235 |
| **Demographics** | | | |
| Age (average; SD) | 61.8; 16.9 | 58.2; 18.0 | 62.3; 17 |
| Female sex (n docs; %) | 353 (51%) | 343 (52%) | 2,128 (50%) |
| **Main document types (n docs, %)** | | | |
| Outpatient report | 340 (49%) | 196 (30%) | 2,184 (52%) |
| Inpatient report | 89 (13%) | 140 (21%) | 483 (11%) |
| Discharge report | 263 (38%) | 322 (49%) | 1,519 (36%) |
| Referral letters | 4 (1%) | 0 (0%) | 28 (1%) |
| **Language distribution** (n docs; n words, % of total) | | | |
| English | 100; 44,582 (10.4%) | 94; 56,098 (14.2%) | 605; 409,422 (13.0%) |
| Spanish | 100; 50,824 (11.8%) | 94; 59,928 (15.2%) | 605; 507,534 (16.2%) |
| Italian | 99; 34,727 (8.1%) | 94; 59,933 (15.2%) | 605; 457,349 (14.6%) |
| Dutch | 100; 40,694 (9.5%) | 94; 54,241 (13.8%) | 605; 424,572 (13.5%) |
| Romanian | 100; 120,048 (27.9%) | 94; 59,959 (15.2%) | 605; 443,624 (14.1%) |
| Czech | 100; 108,865 (25.3%) | 94; 53,707 (13.6%) | 605; 451,895 (14.4%) |
| Swedish | 100; 30,715 (7.1%) | 94; 50,581 (12.8%) | 605; 447,263 (14.2%) |
| **Main heart diseases or conditions (**n docs; %) | | | |
| Heart failure | 188 (27%) | 399 (61%) | 917 (22%) |
| Ischemic heart disease | 199 (28%) | 287 (44%) | 1,106 (26%) |
| Valve diseases | 79 (11%) | 238 (36%) | 315 (7%) |
| Arrhythmias | 66 (9%) | 168 (26%) | 294 (7%) |
| Atrial fibrillation | 60 (9%) | 154 (23%) | 266 (6%) |
| Cardiac arrest | 8 (1%) | 21 (3%) | 35 (1%) |

Overall, the corpus has a balanced sex distribution, with female cases accounting for 51% of the documents and the mean age is 61.8 ± 16.9 years. The most prevalent main cardiac conditions are ischemic heart disease (n=199 documents; 28%), heart failure (n=188; 27%), valvular heart disease (n=79; 11%), arrhythmias (66; 9%), and atrial fibrillation (n=60; 9%), among others, reflecting the most common cardiac diseases seen in cardiology practice. Furthermore, the corpus includes many other common reasons for cardiology consultation such as syncope, chest pain, palpitations and dyspnea (Supplementary Material). Importantly, the documents contain additional cardiovascular diagnoses; for instance, a patient with a primary diagnosis of heart failure may also have hypertension and a history of myocardial infarction. Less frequent diagnoses, such as postpartum cardiomyopathy or myocarditis, are included in the category *Other*.

## ParaCliTe

Applying the methodology described in the Methods Section, resulted in the inclusion of 94 unique adult patient cases, each available in all seven languages, yielding a total of 658 documents (Table 1, Figure 2, Supplementary Material). Each case is linked to structured metadata, enabling stratified evaluation by language, condition, encounter type, or demographic information, without requiring post hoc inference from the text. The final corpus contains 394,447 words, with a deliberately narrow range of approximately 50 to 60 thousand words per language (Table 1, Figure 2). The consistent word counts across languages indicate that the translation process preserved a comparable content volume across language versions, which is important for fair evaluation in multilingual settings.

The mean age of the patients is 58 and standard deviation is 18 years, with a distribution skewed toward older groups, which aligns with the typical epidemiology of cardiovascular disease, while 14 (15%, 98 documents) of these fictional patients were smokers. The ParaCliTe corpus, similarly to CliTe, is also sex-balanced, both in terms of number of documents (343 documents for female patients and 315 for male patients) and number of words (189,655 words are associated with female cases and 204,792 words with male cases, representing 48.1% and 51.9% of the total words, respectively). ParaCliTe covers multiple encounter types, with discharge reports (n=322; 49%), inpatient (n=140; 21%) and outpatient reports (n=196; 30%) as the main categories. This mix supports evaluation across settings that differ in style and information density (e.g., discharge from hospital vs. follow-up visits). In terms of main cardiac diagnoses, the corpus emphasizes common conditions, including heart failure (n=399 documents per language; 61%), ischaemic heart disease (n=287; 44%), valvular heart disease (n=238; 36%), arrhythmias (n=168; 26%) and atrial fibrillation (n=154, 23%), and relevant comorbidities like hypertension (n=266; 40%) and diabetes (n=126; 19%). The main reasons for consultation were dyspnoea (n=301; 46%), chest pain (n=245; 37%), oedema (n=112; 17%), and palpitations (n=70; 11%) (Table 2). Beyond diagnostic labels and comorbidities, the texts contain relevant symptoms, risk factors, treatments and medical history characteristics. For example, texts include medication patterns, procedures, and mentions of implantable devices such as stents, pacemakers, and valves, including explicit statements of absence when clinically appropriate, thus supporting a broad range of downstream clinical NLP tasks.

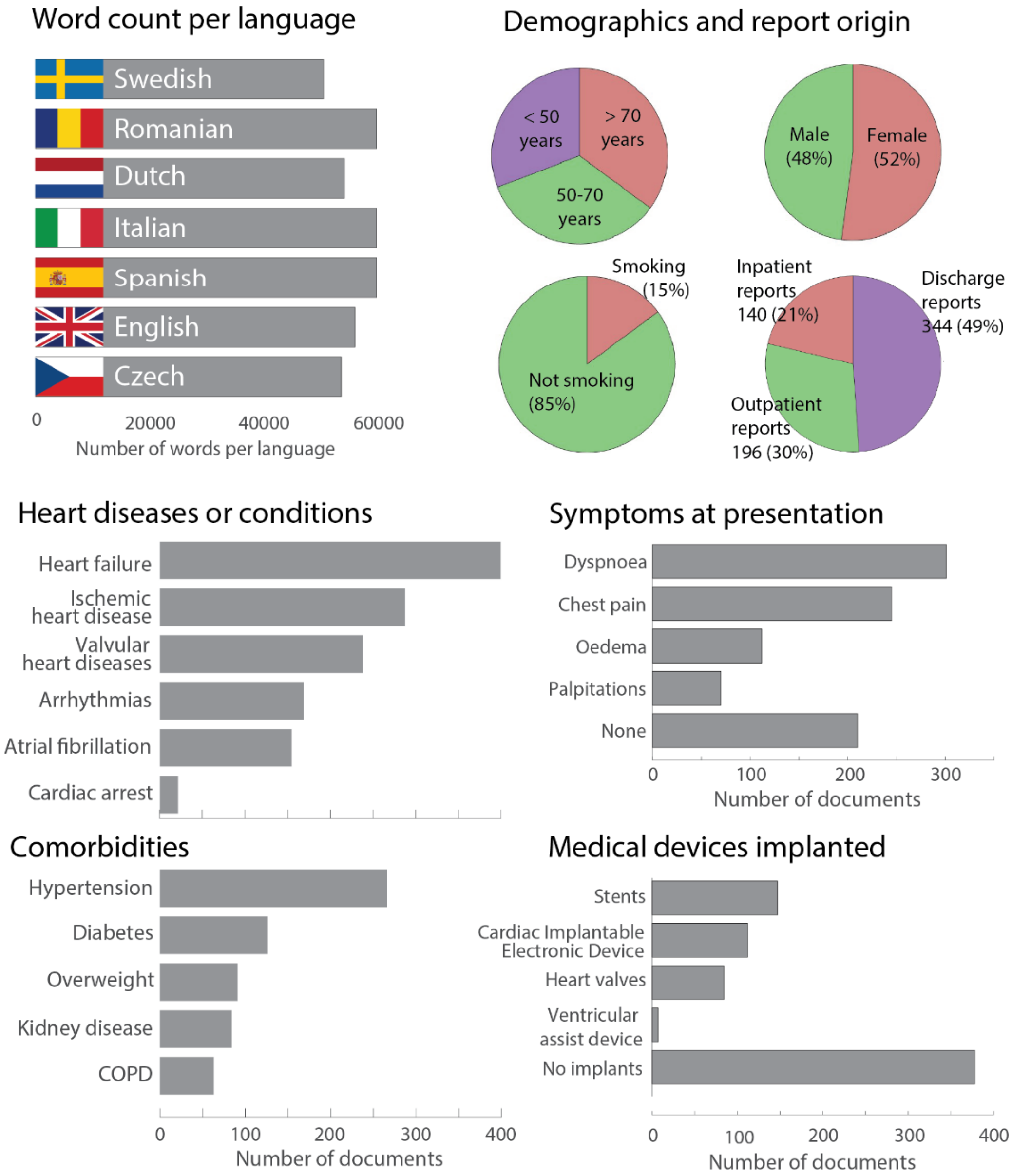

**Figure 2**. Visual representation of the main characteristics of ParaCliTe. COPD = chronic obstructive pulmonary disease.

**Table 2**: Comorbidities and Symptoms present in the ParaCliTe corpus

| **Main comorbidities** | |
|---|---|
| Smoking (n, %) | 98 (15%) |
| Hypertension (n doc, %) | 266 (40%) |
| Diabetes (n doc, %) | 126 (19%) |
| Overweight or obesity (n doc, %) | 91 (14%) |
| Kidney disease (n doc, %) | 84 (13%) |
| COPD (n doc, %) | 63 (10%) |
| **Symptoms** | |
| Dyspnoea (n doc, %) | 301 (46%) |
| Oedema (n doc,%) | 112 (17%) |
| Chest pain (n doc, %) | 245 (37%) |
| Palpitations (n doc, %) | 70 (11%) |
| None (n doc, %) | 210 (32%) |

## Utility Assessment Experiments

In order to explore the utility of the generated resources, we provide initial empirical results to demonstrate the potential of ParaCliTe and offer evidence of the limitations of current baseline models when applied to cardiology texts.

We first evaluated the dedicated MT architectures, utilizing two models of the NLLB family (600M-distilled and 3B parameter variants). As shown in Table 3, the baseline zero-shot performance of these models indicates their limitations in accurately translating the specific nuances of cardiology-related clinical content. However, fine-tuning on the ParaCliTe training splits yielded substantial, double-digit percentage improvements across the vast majority of metrics. Notably, the smaller NLLB-600M-distilled model experienced a dramatic 40.8% relative increase in BLEU score and a 10.8% increase in COMET. This steep increase in performance underscores the potential of ParaCliTe to mitigate the lack of the specific linguistic mappings required for this domain in off-the-shelf models. To assess the utility of the dataset for LLMs without the confounding variables of zero-shot generative formatting, we computed the PPL on the ParaCliTe test set before and after fine-tuning the models. Table 3 illustrates a significant reduction in PPL across all tested LLM architectures, ranging from 1B to 12B parameters. For instance, Llama-3.2-1B saw its PPL more than halve (from 6.13 to 2.61), while models like Qwen3 8B and Gemma 2 9B reached highly confident PPL scores of 1.57 and 1.63, respectively. High baseline PPL confirms that the linguistic distribution of ParaCliTe is unfamiliar to these foundational models. The sharp drop post-fine-tuning provides intrinsic evidence that the dataset successfully updates the models' internal representations, teaching them the target domain's semantic and syntactic rules.

Once the LLMs were fine-tuned, they learned to reliably follow the instructional format required to output clean translations. This permitted the evaluation of all fine-tuned models, both dedicated MT and LLMs, using standard generative metrics to establish the current state-of-the-art benchmarks on the ParaCliTe dataset. As detailed in Table 3, the dedicated MT models remain highly competitive and efficient, and some fine-tuned LLMs are capable of matching or slightly exceeding their performance. Gemma 2 9B achieved the highest overall generative metrics (COMET 0.85, BLEU 0.39, chrF++ 0.60), closely followed by Qwen3 8B.

**Table 3**: Performance changes comparing pre vs. post fine-tuning.

| Model | Fine-tuned | COMET | COMET-Kiwi | BLEU | chrf++ | PPL |
|---|---|---|---|---|---|---|
| Machine translation models | | | | | | |
| NLLB-600M-distilled | NO | 0.76 | 0.69 | 0.26 | 0.49 | - |
| | YES | 0.84 (+10.8%) | 0.73 (+5.8%) | 0.37 (+40.8%) | 0.59 (+20.6%) | - |
| NLLB-3B | NO | 0.77 | 0.70 | 0.28 | 0.50 | - |
| | YES | 0.84 (+9.6%) | 0.73 (+4.3%) | 0.36 (+28.8%) | 0.59 (+17.5%) | - |
| Large language models | | | | | | |
| Llama-3.2-1B | NO | - | - | - | - | 6.13 |
| | YES | 0.80 | 0.70 | 0.25 | 0.48 | 2.61 (-57.4%) |
| Llama-3.1-8B | NO | - | - | - | - | 2.77 |
| | YES | 0.80 | 0.70 | 0.28 | 0.51 | 1.63 (-41.2%) |
| Gemma 2 9B | NO | - | - | - | - | 2.51 |
| | YES | **0.85** | **0.75** | **0.39** | **0.60** | **1.63** (-35.1%) |
| Mistral v0.3 7B | NO | - | - | - | - | 4.13 |
| | YES | 0.79 | 0.68 | 0.25 | 0.49 | 1.94 (-53.0%) |
| Qwen3 8B | NO | - | - | - | - | 4.50 |
| | YES | 0.84 | 0.74 | 0.36 | 0.57 | 1.57 (-65.1%) |
| Mistral-Nemo-12B | NO | - | | - | - | 4.55 |
| | YES | 0.79 | 0.68 | 0.26 | 0.48 | 1.81 (-60.2%) |

Bold identify the best-performing model. PPL = perplexity.

## ParaCliTe Silver

We extended our evaluation to the ParaCliTe Silver dataset using the highest-performing automatic MT model, Gemma 2 9B (Table 1). As the ParaCliTe Silver corpus lacks human reference translations, reference-based metrics like COMET or BLEU cannot be applied. Instead, we evaluated the translations using COMET-Kiwi, a reference-free quality estimation model for MT. To verify the metric's applicability to this domain, we first calculated COMET-Kiwi on the gold ParaCliTe test set and observed a positive correlation with the reference-based COMET scores. For Gemma 2 9B, the COMET-Kiwi score was 0.75 on the gold test set and 0.72 on the silver corpus. The small difference between these scores suggests that the performance observed on the test set is maintained on the larger-scale ParaCliTe Silver corpus.

The automatic translation resulted in a corpus containing 4,235 documents and a total of 3.1 million words (Table 1). As CliTe and ParaCliTe, ParaCliTe Silver is sex-balanced, with 2,128 documents (50%) for female patients, well distributed across the 7 languages, and covers the most common main cardiac diagnoses and document types.

# Discussion

This study represents one of the first attempts to address the current bottleneck and limitations in access to comparable multilingual cardiology reports. In particular, it introduces an openly-available framework defining clear data generation principles to establish credible and consistent artificial clinical text datasets that avoid harmful biases, reflect the true diversity of the local patient population, and avoid privacy concerns. While developed for the cardiology domain, such a resource could be adapted to other medical specialties with minor adaptations. The proposed strategy was exploited to create a diverse, high-quality multilingual cardiology clinical corpus covering seven European languages. To the best of our knowledge, this work is the first effort to create multilingual cardiology clinical report datasets, addressing the need for accessible benchmarking resources, which are essential for adapting and developing AI and more specifically NLP and LLM solutions. To date, such resources are scarce due to privacy concerns and the limited public sharing of real-world clinical texts. Indeed, most clinical NLP research is based on English-language clinical data from a limited number of clinical sites in the United States, such as MIMIC-IV [14]. This highlights the need for European and multilingual clinical data resources and methods that support clinical data processing beyond English to enhance generalizability of language models and to assist the developments required for the effective implementation of the European Health Data Space.

In this respect, this study releases three multilingual corpora, namely CliTe, ParaCliTe and ParaCliTe Silver, to provide the community with resources for advancing and validating NLP solutions specifically tailored to clinical cardiology. A key contribution of the presented work is ParaCliTe, which was developed through a careful multi-stage manual translation process, including feedback loop with cardiologists, resulting in a gold-standard reference dataset for training and evaluating NLP models and machine translation systems adapted to cardiology. The artificial clinical reports included in these corpora are designed to reflect patient characteristics, local clinical practices, and the expertise of large European hospitals. They incorporate structural nuances and regional specificities to capture real-world clinical settings while minimizing the risk of propagating disparities across patient groups.

Additionally, the public release of these corpora will help foster research and support the development and implementation of clinical NLP and AI resources for languages beyond English, with a particular focus on cardiovascular disease patients. This is crucial since, in contrast with clinical research, provision of care is performed in local languages. In this context, a previous work reported the generation of artificial radiology reports spanning 21 countries and 13 languages [35], while another reported artificial clinical reports from 18 medical specialties in French [36]. However, no previous multilanguage corpus has been generated for cardiology, a key limitation of the domain now covered by the CliTe and ParaCliTe corpora. The presented results from the utility assessment suggest that ParaCliTe not only serve as an evaluation or benchmark resource but can also be useful for the fine-tuning of both MT or general-purpose LLMs to improve their performance in the cardiology domain. The results in this study underline the potential of such corpora, and the urgency to extend the number of languages and clinical specialties covered by these resources. In this respect, CliTe and ParaCliTe Silver are currently being expanded to include additional content from Europe (Czech Republic, in Czech), Latin America (Peru, in Latin American Spanish) and Africa (Tanzania, in English).

The importance of these efforts is evident as sharing real-world health data is essential for cross-border healthcare and improved patient outcomes, yet privacy constraints often limit such data sharing. Although multilingual, multisite clinical datasets are relatively costly and time-intensive to develop, artificially generated datasets reflecting real clinical narratives are persistent, accessible and shareable resources that can be further extended, and updated, thereby supporting the sustainable development of clinical language technology solutions for cardiovascular content processing.

Nevertheless, the presented work and its results have several limitations that should be considered and addressed in future research. Although the presented datasets are already valuable for the domain adaptation of clinical NLP and MT systems, larger and more heterogeneous corpora of clinical reports could further enhance linguistic and lexical diversity. For instance, the presented dataset could be improved by providing better coverage of less common patient characteristics and could also be extended to include additional languages that are not currently represented. In light of recent advances in generative clinical LLMs, prompt engineering–based data augmentation techniques could be applied using CliTe and ParaCliTe, together with the corpus construction guidelines, in zero-shot or few-shot settings as initial seed examples. This approach could be used to systematically generate larger collections of clinical reports, incorporate additional clinical features and attributes, expand the content to include additional languages and language varieties, increase the size of the datasets, and ultimately improve downstream model training. It is also important to note that the generation of CliTe and ParaCliTe was carried out by clinical professionals representing the expertise of seven major European hospitals. As a result, the datasets reflect the medical settings, clinical practices, and terminology commonly used in these institutions, which may differ from those found in other clinical environments. Consequently, the datasets may reflect, to some extent, institution-specific clinical practices that are not fully representative of broader clinical settings. Researchers should be mindful of these considerations when generalizing findings from models developed using these resources. Lastly, the datasets were explicitly created using fictional patients for the development of AI tools. As such, their intended use does not include deriving clinical insights or directly developing clinical predictive models for real patient populations, even though the corpus construction guidelines were designed to provide a realistic representation of local clinical settings.

## Conclusions

The presented corpus construction protocol and guidelines, together with the multilingual fictitious patient corpora, constitute a valuable resource for the development and evaluation of advanced cardiology-adapted language models and language technology solutions across seven European languages. By leveraging artificial texts that reflect local clinical structures and content while promoting linguistic inclusivity, these datasets address a key bottleneck for developing AI-based solutions that preserve patient privacy while providing clinically realistic data for research and development.

## Acknowledgements

We would like to thank the professional translators that participated in the production of ParaCliTe.

## Funding

This work has received funding from the European Union's Horizon Europe research and innovation programme under grant agreement 101057849 and 101080430. A.G. received funding from the Instituto de Salud Carlos III (CP24/00121). This publication was made possible through funding from the Sector Plan for Medicine and Health Sciences, supported by the Dutch Ministry of Education, Culture and Science (OCW).

## Disclosure of interest

None

## Data availability statement

The *Guiding Principles for Multilingual Generation of Artificial Clinical Reports* are published in the Zenodo repository (https://zenodo.org/records/21220631), while CliTe, ParaCliTe and ParaCliTe Silver can be accessed at the Zenodo repository (https://zenodo.org/records/22641301).

## Notes

DataTools4Heart NLP group: Karim Lekadir (Universitat de Barcelona, Spain), Xenia Puig Bosch (Universitat de Barcelona, Spain), Anoop Shah (University College London Hospital, UK), Cristian Izquierdo (Universitat de Barcelona, Spain), Folkert Asselbergs (Amsterdam University Medical , The Netherlands), Saskia Haitjema (University Medical Center Utrecht, The Netherlands), Jacopo Lenkowicz (Fondazione Policlinico Universitario A. Gemelli IRCCS, Italy), Roman Panovsky (St. Anne's University Hospital Brno, Czech Republic), George Rayers (University College London Hospital, UK), Isaac Harris (University College London Hospital, UK), George Katis (University College London Hospital, UK), Keya Pindolia (University College London Hospital, UK), Baker Kirresh (University College London Hospital, UK), Shivam Mehta (University College London Hospital, UK), Ruman Basra (University College London Hospital), Seshasai Narimetla (University College London Hospital, UK)

# Supplementary Material

## Experimental Setup and Fine-Tuning Procedures

**Hardware and Computational Resources:** All fine-tuning and inference procedures for both the machine translation (MT) models and the large language models (LLMs) were conducted using a single NVIDIA GeForce RTX 3090 GPU equipped with 24GB of VRAM. This specific hardware constraint was intentionally selected to ensure that the proposed translation and data generation pipelines remain accessible and fully deployable within standard clinical sites, which typically lack access to high-end, data-center-grade computing clusters.

**MT Models Setup:** For the dedicated MT models, NLLB-600M-distilled and NLLB-3B, fine-tuning was framed as a standard sequence-to-sequence generation task. The NLLB-600M-distilled model was subjected to full parameter fine-tuning. We utilized the AdamW optimizer with a learning rate of 5e-5, a weight decay of 0.01, and a linear learning rate scheduler with a 10% warmup ratio. To maintain a functional batch size within the GPU's memory limits, we employed a per-device micro-batch size of 4 paired with gradient accumulation steps set to 8. For the larger NLLB-3B model, memory constraints necessitated the use of Low-Rank Adaptation (LoRA) [1] applied to the attention modules, utilizing a rank of 16 and an alpha of 32, while keeping the base model weights frozen in 8-bit precision. Both models were trained for 5 epochs with early stopping configured to halt training if the validation loss did not improve over 3 consecutive evaluations.

**LLMs Setup:** Fine-tuning the suite of LLMs (Llama 3.2 1B, Llama 3.1 8B, Mistral v0.3 7B, Qwen3 8B, Gemma 2 9B, and Mistral-Nemo-12B) on a single RTX 3090 required aggressive memory optimization. We employed QLoRA (Quantized Low-Rank Adaptation) [2], loading the base models in 4-bit NormalFloat (NF4) precision with double quantization and a 16-bit brain floating-point (bfloat16) compute data type. Gradient checkpointing was enabled across all LLM experiments to further reduce VRAM utilization.

The LoRA adapters were applied to all linear layers (including query, key, value, output, gate, up, and down projections) to maximize the trainable parameter surface for domain adaptation. The LoRA configuration utilized a rank of 32, a scaling alpha of 64, and a dropout rate of 0.05. Training was executed using the paged_adamw_8bit optimizer to prevent memory spikes during state updates. The learning rate was set to 2e-4, regulated by a cosine annealing learning rate scheduler with a 5% warmup period. We used a micro-batch size of 2 and accumulated gradients over 16 steps to simulate a larger effective batch size. All LLMs were fine-tuned for 3 epochs using a standard causal language modeling objective on the packed conversational prompt formats containing the source clinical text and the target ParaCliTe translations.

**Evaluation Metrics Setup**: Automated evaluation was conducted using standard pre-trained models from the Unbabel framework. Specifically, reference-based COMET scores were computed using the Unbabel/wmt22-comet-da model, and reference-free COMET-Kiwi quality estimation was conducted using the Unbabel/wmt22-cometkiwi-da model.

## ParaCliTe corpora metadata

List of metadata accompanying the ParaCliTe corpora.

**Type of document**

- Outpatient report
- Hospital admission/inpatient report
- Emergency Department report

**Symptoms**

- Dyspnoea/Orthopnea/Shortness of breath
- Oedema
- Chest pain
- Palpitations
- NYHA

**Heart Disorders**

- Arrhythmia (not atrial fibrillation)
- Atrial fibrillation
- Heart failure
- Heart valve disease
- Hypertension
- Myocardial infarction/Ischaemic heart disease
- Cardiac arrest
- Cardiac devices, including:
    - Cardiac implantable electronic devices (Pacemakers/CRT-D/P/ICD)
    - Heart valves
    - Stents
    - Ventricular assist devices (LVAD/IABP)

**Comorbidities**

- Overweight/Obesity/BMI>25
- Diabetes
- Kidney disease
- Chronic obstructive pulmonary disease
- Smoker
- Reduced mobility

## CliTe Main Diseases

**In order or decreasing representativity**:

- Ischemic Heart Disease
- Heart failure
- Heart valve disease
- Arrythmia
- Dilated cardiomyopathy
- Atrial fibrillation
- Syncope
- Lung disease
- Chest pain
- Obstructive cardiomyopathy
- Pericardial disorder
- Myocarditis
- Palpitations
- Hypertension
- Cardiac arrest
- Congenital heart disease
- Endocarditis
- Heart disease (unspecified)
- Dyspnea
- Cardiogenic shock
- Non-ischemic cardiomyopathy

**Conditions with less than 3 fictional patients**:

- Aortic dissection
- Cardiac amyloidosis
- Cerebrovascular disease
- Chemotoxic-induced cardiopathy
- Dizziness
- Fatigue
- Left Ventricular Hypertrophy
- Postpartum cardiomyopathy
- Cough
- Heart murmur
- Hypercholesterolemia
- Peripheral artery disease
- Sepsis
- Shock
- Stress
- Vertigo
- Other unspecified conditions

## Supplementary material references